\documentclass[11pt,a4paper]{article}
\usepackage{emnlp2020}
\usepackage{times}
\usepackage{latexsym}

\usepackage{microtype}
\usepackage{times}  
\usepackage{helvet}  
\usepackage{courier}  
\usepackage{url}  
\usepackage{graphicx}  
\usepackage{amsfonts}       
\usepackage{amsmath}
\usepackage{amssymb}
\usepackage{bbm}
\usepackage{amsthm}
\usepackage{pifont}
\usepackage{bm}
\usepackage{booktabs}       
\usepackage{soul}
\usepackage{multirow}
\usepackage{subcaption}

\usepackage[ruled,vlined]{algorithm2e}
\usepackage{makecell}
\newcommand{\cmark}{\ding{51}}%
\newcommand{\xmark}{\ding{55}}%

\aclfinalcopy 
\def\aclpaperid{3368} 

\title{PrivNet: Safeguarding Private Attributes in Transfer Learning for Recommendation}

\author{Guangneng Hu\\
  HKUST / Hong Kong, China \\
  \texttt{njuhgn@gmail.com} \\\And
  Qiang Yang \\
  HKUST / Hong Kong, China \\
  \texttt{qyang@cse.ust.hk} \\}

\begin{document}
\maketitle

\begin{abstract}
Transfer learning is an effective technique to improve a target recommender system with the knowledge from a source domain. Existing research focuses on the recommendation performance of the target domain while ignores the privacy leakage of the source domain. The transferred knowledge, however, may unintendedly leak private information of the source domain. For example, an attacker can accurately infer user demographics from their historical purchase provided by a source domain data owner. This paper addresses the above privacy-preserving issue by learning a privacy-aware neural representation by improving target performance while protecting source privacy. The key idea is to simulate the attacks during the training for protecting unseen users' privacy in the future, modeled by an adversarial game, so that the transfer learning model becomes robust to attacks. Experiments show that the proposed PrivNet model can successfully disentangle the knowledge benefitting the transfer from leaking the privacy.
\end{abstract}

\section{Introduction}

Recommender systems (RSs) are widely used in everyday life ranging from Amazon products~\cite{zhou2018deep,wan2020addressing} and YouTube videos~\cite{gao2010vlogging,cheng2016wide} to Twitter microblogs~\cite{huang2016hashtag} and news feeds~\cite{wang2018dkn,ma2019news2vec}. RSs estimate user preferences on items from their historical interactions. RSs, however, cannot learn a reliable preference model if there are too few interactions in the case of new users and items, i.e., suffering from the data sparsity issues.

Transfer learning is an effective technique for alleviating the issues of data sparsity by exploiting the knowledge from related domains~\cite{pan2010transfer,liu2018transferable}. We may infer user preferences on videos from their Tweet texts~\cite{huang2016transferring}, from movies to books~\cite{li2009can}, and from news to apps~\cite{hu2018conet,TMH}. These behaviors across domains are different views of the same user and may be driven by some inherent user interests~\cite{elkahky2015multi}.

There is a privacy concern when the source domain shares their data to the target domain due to the ever-increasing user data abuse and privacy regulations~\cite{ramakrishnan2001privacy,yang2019federated}. Private information contains those attributes that users do not want to disclose, such as gender and age~\cite{jia2018attriguard}. They can be used to train better recommendation models by alleviating the data sparsity issues to build better user profiles~\cite{zhao2014we,cheng2016wide}. Previous work~\cite{weinsberg2012blurme,beigi2020privacy} shows that an attacker can accurately infer a user's gender, age, and occupation from their ratings, recommendation results, and a small amount of users who reveal their demographics.

\begin{figure}
\centering
\begin{subfigure}{.23\textwidth}
  \includegraphics[height=1.3in,width=3.8cm]{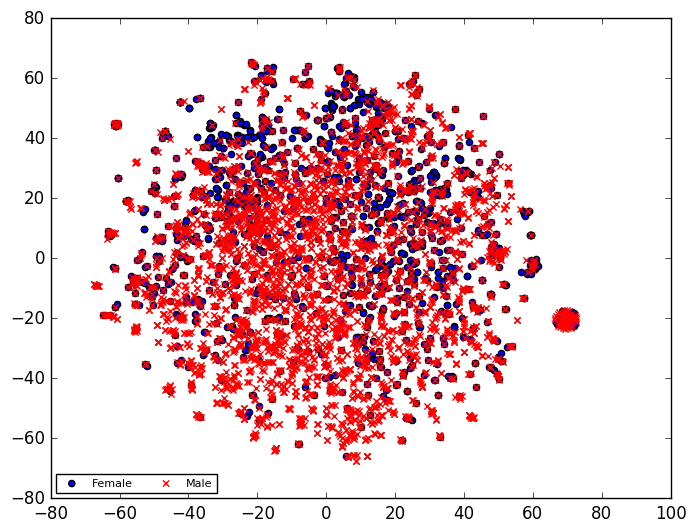}
\end{subfigure}
\begin{subfigure}{.22\textwidth}
  \includegraphics[height=1.3in,width=3.8cm]{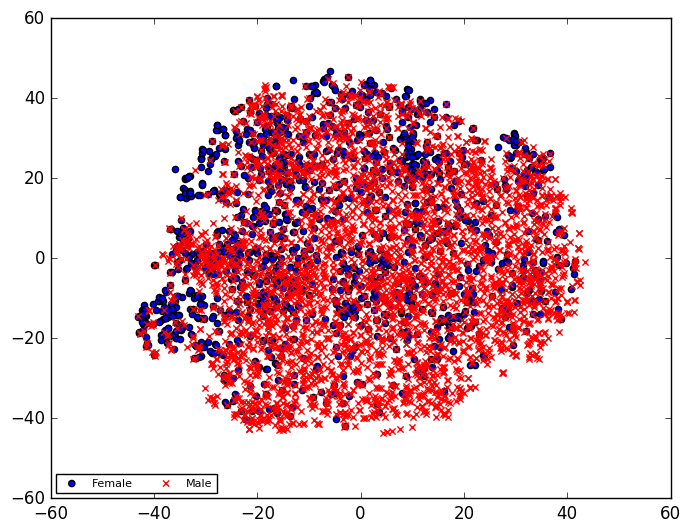}
\end{subfigure}
\caption{t-SNE projection of transferred representations of users with (left) and without (right) training of PrivNet on the MovieLens-Gender dataset. (see Section~\ref{paper:result-clustering} for details)}
\label{fig:tsne}
\end{figure}

A technical challenge for protecting user privacy in transfer learning is that the transferred knowledge has dual roles: usefulness to improve target recommendation and uselessness to infer source user privacy. In this work, we propose a novel model (PrivNet) to achieve the two goals by learning privacy-aware transferable knowledge such that it is useful for improving recommendation performance in the target domain while it is useless to infer private information of the source domain. The key idea is to simulate the attack during the training for protecting unseen users' privacy in the future. The privacy attacker and the recommender are naturally modeled by an adversarial learning game. The main contributions are two-fold:
\begin{itemize}
\item PrivNet is the first to address the privacy protection issues, i.e., protecting source user private attributes while improving the target performance, during the knowledge transfer in neural recommendation.
\item PrivNet achieves a good tradeoff between the utility and privacy of the source information through evaluation on real-world datasets by comparing with strategies of adding noise (i.e., differential privacy) and perturbing ratings.
\end{itemize}

\section{Related Work}

\subsection{Transfer learning in recommendation}
Transfer learning in recommendation~\cite{cantador2015cross} is an effective technique to alleviate the data sparsity issue in one domain by exploiting the knowledge from other domains. Typical methods apply matrix factorization~\cite{singh2008relational,pan2010transfer,Yang2017MUB} and representation learning~\cite{zhang2016collaborative,man2017cross,yang2017bridging,gao2019cross,Ma2019PI} on each domain and share the user (item) factors, or learn a cluster level rating pattern~\cite{li2009can,Yuan2019darec}. Transfer learning is to improve the target performance by exploiting knowledge from auxiliary domains~\cite{pan2009survey,elkahky2015multi,zhang2017survey,Chen2019EAT,gao2019neural}. One transfer strategy (two-stage) is to initialize a target network with transferred representations from a pre-trained source network~\cite{oquab2014learning,yosinski2014transferable}. Another transfer strategy (end-to-end) is to transfer knowledge in a mutual way such that the source and target networks benefit from each other during the training, with examples including the cross-stitch networks~\cite{misra2016cross} and collaborative cross networks~\cite{hu2018conet}. These transfer learning methods have access to the input or representations from source domain. Therefore, it raises a concern on privacy leaks and provides an attack possibility during knowledge transfer.

\subsection{Privacy-preserving techniques}
Existing privacy-preserving techniques mainly belong to three research threads. One thread adds noise (e.g., differential privacy~\cite{dwork2006calibrating}) to the released data or the output of recommender systems~\cite{mcsherry2009differentially,jia2018attriguard,meng2018personalized,wang2018differentially,wang2020pifferentially}. One thread perturbs user profiles such as adding (or deleting/changing) dummy items to the user history so that it hides the user's actual ratings~\cite{polat2003privacy,weinsberg2012blurme}.  Adding noise and perturbing ratings may still suffer from privacy inference attacks when the attacker can successfully distinguish the true profiles from the noisy/perturbed ones. Furthermore, they may degrade performance since data is corrupted. Another thread uses adversary loss~\cite{Resheff19Privacy,beigi2020privacy} to formulate the privacy attacker and the recommender system as an adversarial learning problem. However, they face the data sparsity issues. A recent work~\cite{ravfogel2020null} trains linear classifiers to predict a protected attribute and then remove it by projecting the representation on its null-space. Some other work uses encryption and federated learning so as to protect the personal data without affecting performance~\cite{nikolaenko2013privacy,chen2018federated,ijcai2019novel}. They suffer from efficiency and scalability due to high cost of computation and communication.

\section{Problem Statement}\label{paper:statement}

We have two domains, a source domain $S$ and a target domain $T$. User sets in two domains are shared, denoted by $\mathcal{U}$ (of size $m=|\mathcal{U}|$). Denote item sets in two domains by $\mathcal{I}_S$ and $\mathcal{I}_T$ (of size $n_S=|\mathcal{I}_S|$ and $n_T=|\mathcal{I}_T|$), respectively. For the target domain, a binary matrix $\bm{R}_T \in \mathbb{R}^{m \times n_T}$ describes the user-item interactions, where the entry $r_{ui}  \in \{0,1\}$ equals 1 if user $u$ has an interaction with item $i$ and 0 otherwise. Similarly, for the source domain, we have $\bm{R}_S \in \mathbb{R}^{m \times n_S}$ and the entry $r_{uj} \in \{0,1\}$. We reserve $i$ and $j$ to index the target and source items, respectively. Let $\bm{Y}^p \in \mathbb{R}^{m \times c_p}$ denote the $p$-th user private attribute (e.g., $p$=`Gender') matrix where each entry $y_{u,p}$ is the value of the $p$-th private information for user $u$ (e.g., $y_{u,p}$=`Male') and there are $c_p$ choices. Denote all $n$ private attributes data by $\bm{Y} = \{\bm{Y}^p\}_{p=1}^n$ (e.g., Gender, Age). We can define the problem as follows:

{\scshape Problem:} Privacy-aware transfer learning in recommendation.

{\scshape Input:} $\bm{R}_T, \bm{R}_S, \bm{Y}$.

{\scshape Output:} Generate a ranked list of items for users in the target domain.

{\scshape Require:} An attacker is difficult to infer the source user private attributes from the knowledge transferred to the target domain.

{\scshape Assumption:} Some users $\mathcal{U}^{pub} \subset \mathcal{U}$ share their private information with the public profile.

\begin{figure}
\centering
\includegraphics[height=2.2in,width=7.8cm]{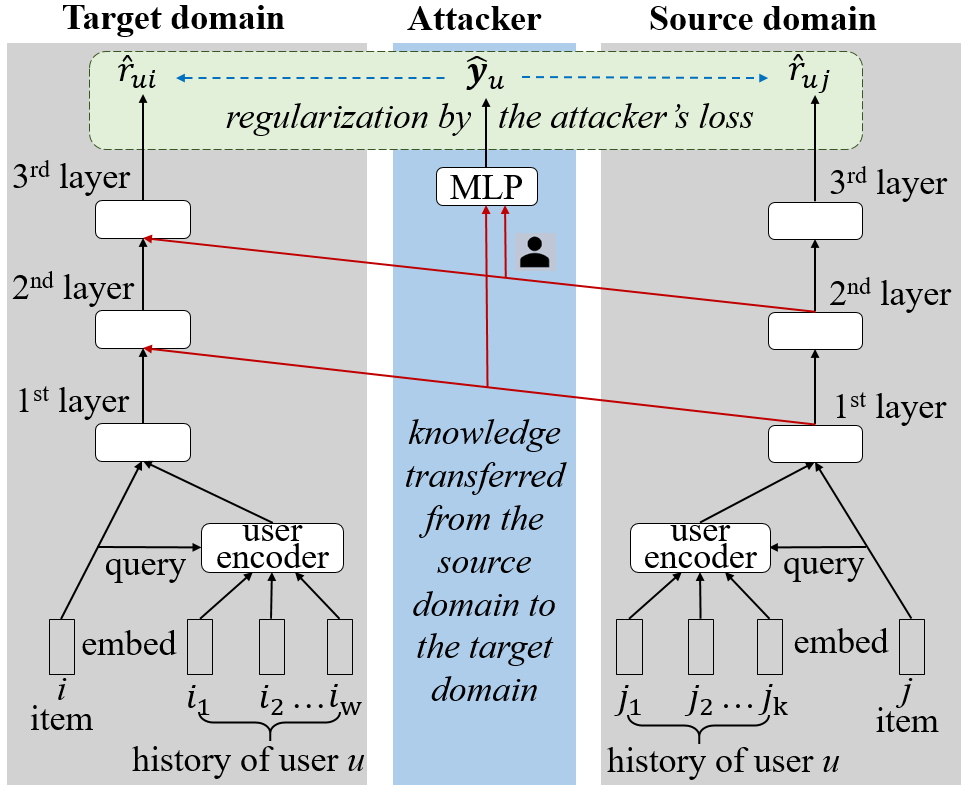}
\caption{Architecture of PrivNet (a version of three layers). It has two components: the recommender and privacy attacker. The recommender (the left \& right parts, see Section~\ref{paper:rec}) is a representation-based transfer learning model where the red arrows indicate the representations transferred from the source domain to the target domain in a multilayer way. The privacy attacker (the middle part, see Section~\ref{paper:attacker}) marked by an avatar infers user privacy from the transferred representations. PrivNet (see Section~\ref{paper:privnet}) exploits the knowledge from the source domain with regularization from the adversary loss of the attacker indicated by the dotted box.}
\label{fig:architecture}
\end{figure}

\section{The Proposed Framework}\label{paper:proposed-model}

The architecture of PrivNet is shown in Figure~\ref{fig:architecture}. It has two components, a recommender and an attacker. We introduce the recommender (Section~\ref{paper:rec}) and present an attack against it (Section~\ref{paper:attacker}). We propose PrivNet to protect source user privacy during the knowledge transfer (Section~\ref{paper:privnet}).

\subsection{Recommender}\label{paper:rec}

In this section, we introduce a novel transfer-learning recommender which has three parts, a source network for the source domain, a target network for the target domain, and a knowledge transfer unit between the two domains.

{\bf Target network} The input is a pair of (user, item) and the output is their matching degree. The user is represented by their $w$-sized historical items $[i_1,...,i_w]$. First, an item embedding matrix $\bm{A}_T$ projects the discrete item indices to the $d$-dimensional continuous representations: $\bm{x}_i$ and $\bm{x}_{i_*}$ where $* \in [1,2,...,w]$. Second, the user representation $\bm{x}_u$ is computed by the user encoder module based on an attention mechanism by querying their historical items with the predicted item: $\bm{x}_u = \sum\nolimits_{i_*} \alpha_{i_*} \bm{x}_{i_*},$ where $\alpha_{i_*} = \bm{x}_i^T \bm{x}_{i_*}$ (normalized: $\sum \alpha_{i_*} = 1$). Third, a multilayer perceptron (MLP) $f_T$ parameterized by $\phi_T$ is used to compute target preference score (the notation $[\cdot,\cdot]$ denotes concatenation):
\begin{equation*}\label{eq:neural-cf-target}
\hat{r}_{ui} = P(r_{ui}|u,i;\theta_T) = f_T([\bm{x}_u,\bm{x}_i]),
\end{equation*}
where $\theta_T = \{\bm{A}_T, \phi_T\}$ is the model parameter.


{\bf Source network} Similar to the three-step computing process in the target network, we compute the source preference score by: $\hat{r}_{uj} = P(r_{uj}|u,j;\theta_S) = f_S([\bm{x}_u,\bm{x}_j])$ where $\theta_S = \{\bm{A}_S, \phi_S\}$ is the model parameter with item embedding matrix $\bm{A}_S$ and multilayer perceptron $\phi_S$.

{\bf Transfer unit} The transfer unit implements the knowledge transfer from the source to the target domain. Since typical neural networks have more than one layer, say $L$, the representations are transferred in a multilayer way. Let $\bm{x}_{u|\#}^{\ell}$ where $\# \in \{S,T\}$ be user $u$'s source/target representation in the $\ell$-th layer ($\ell=1,2,...,L-1$) where $\bm{x}_{u|S}^{1} = [\bm{x}_u,\bm{x}_j]$ and $\bm{x}_{u|T}^{1} = [\bm{x}_u,\bm{x}_i]$. The transferred representation is computed by projecting the source representation to the space of target representations with a translation matrix $\bm{H}^{\ell}$:
\begin{equation}\label{eq:trans-layer}
\bm{x}_{u|trans}^{\ell} = \bm{H}^{\ell} \bm{x}_{u|S}^{\ell},
\end{equation}

With the knowledge from the source domain, the target network learns a linear combination of the two input activations from both networks and then feeds these combinations as input to the successive layer's filter. In detail, the $(\ell+1)$-th layer's input of the target network is computed by: $\bm{W}_T^{\ell} \bm{x}_{u|T}^{\ell} + \bm{x}_{u|trans}^{\ell}$ where $\bm{W}_T^{\ell}$ is the connection weight matrix in the $\ell$-th layer of the target network. The total transferred knowledge is concatenated by all layers's representations:
\begin{equation}\label{eq:trans}
\bm{x}_{u|trans} = [\bm{x}_{u|trans}^{\ell}]_{\ell=1}^{L-1}.
\end{equation}

{\bf Objective} The recommender minimizes the negative logarithm likelihood:
\begin{multline}\label{eq:loss-rec-vanilla}
\mathcal{L}(\theta) = -\sum\nolimits_{D_T} \log P(r_{ui}|u,i;\theta_T) \\ - \sum\nolimits_{D_S} \log P(r_{uj}|u,j;\theta_S),
\end{multline}
where $\theta = \{\theta_T, \theta_S, \{\bm{H}^{\ell}\}_{\ell=1}^{L-1} \}$, $D_T$ and $D_S$ are target and source training examples, respectively. We introduce how to generate them in Section~\ref{paper:gen-training}.

\subsection{Attacker}\label{paper:attacker}

The recommender can fulfil the Problem 1 (see Section~\ref{paper:statement}) if there is no attacker existing. A challenge for the recommender is that it does not know the attacker models in advance. To address this challenge, we add an attacker component during the training to simulate the attacks for the test. By integrating a simulated attacker into the recommender, it can deal with the unseen attacks in the future. In this section, we introduce an attacker to infer the user private information from the transferred knowledge. In the next Section~\ref{paper:privnet}, we will introduce an adversarial recommender by exploiting the simulated attacker to regularize the recommendation process in order to fool the adversary so that it can protect the privacy of unseen users in the future.

The attacker model predicts the private user attribute from their source representation sent to the target domain:
\begin{equation}\label{eq:attacker-pred}
\hat{y}_{u,p} = P(y_{u,p}|\bm{x}_{u|trans};\theta_p) = f_p(\bm{x}_{u|trans};\theta_p),
\end{equation}
where $\hat{y}_{u,p}$ is the predicted value of user $u$'s $p$-th private attribute and $p=1,...,n$. $f_p$ is the prediction model parameterized by $\theta_p$. Note that, an attacker can use any prediction model and here we use an MLP due to its nonlinearity and generality.

For all $n $ private user attributes, the attacker model minimizes the multitask loss:
\begin{equation}\label{eq:attack-loss}
\mathcal{L}(\Theta) = - \frac{1}{n} \sum\nolimits_p \sum\nolimits_{D_p} \log P(y_{u,p}|\bm{x}_{u|trans};\theta_p),
\end{equation}
where $\Theta = \{\theta_p\}_{p=1}^n$ and $D_p$ is training examples for the $p$-th attribute. We introduce how to generate them in Section~\ref{paper:gen-training}.

\subsection{PrivNet}\label{paper:privnet}

So far, we have introduced a recommender to exploit the knowledge from a source domain and a privacy attacker to infer user private information from the transferred knowledge. To fulfill the Problem 1 in Section~\ref{paper:statement}, we need to achieve two goals: improving the target recommendation and protecting the source privacy. In this section, we propose a novel model (PrivNet) by exploiting the attacker component to regularize the recommender.

Since we have two rival objectives (i.e., target quality and source privacy), we adopt the adversarial learning technique~\cite{goodfellow2014generative} to learn a privacy-aware transfer model. The generator is a privacy attacker which tries to accurately infer the user privacy, while the discriminator is an recommender which learns user preferences and deceives the adversary. The recommender of PrivNet minimizes:
\begin{equation}\label{eq:privnet}
\tilde{\mathcal{L}}(\theta) = \mathcal{L}(\theta) - \lambda \mathcal{L}(\Theta),
\end{equation}
where the hyperparameter $\lambda$ controls the influence from the attacker component. PrivNet seeks to improve the recommendation quality (the first term on the right-hand side) and fools the adversary by maximizing the loss of the adversary (the second term,). The adversary has no control over the transferred knowledge, i.e., $\bm{x}_{u|trans}$. Losses of the two components are interdependent but they optimize their own parameters. PrivNet is a general framework since both the recommender and the attacker can be easily replaced by their variants. PrivNet reduces to privacy-agnostic transfer model when $\lambda=0$.

\subsection{Generating Training Examples}\label{paper:gen-training}

We generate $D_T$ and $D_S$ as follows and take the target domain as an example since the procedure is the same for the source domain. Suppose we have a whole item interaction history for some user $u$, say $[i_1,i_2,...,i_l]$. Then we generate the positive training examples by sliding over the sequence of the history: $D_T^+ = \{([i_w]_{w=1}^{c-1}, i_c): c=2,...,l\}$. We adopt the random negative sampling technique~\cite{pan2008one} to generate the corresponding negative training examples $D_T^- = \{([i_w]_{w=1}^{c-1}, i_c'): i_c' \notin [i_1,i_2,...,i_l]\}$. As the same with~\cite{weinsberg2012blurme,beigi2020privacy}, we assume that some users $\mathcal{U}^{pub} \subset \mathcal{U}$ share their private attributes with the public profile. Then we have the labelled privacy data $D_{priv} = \{D_p\}_{p=1}^n$ where $D_p = \{(u, y_{u,p}): u \in \mathcal{U}^{pub}\}$. 

\begin{algorithm}[t]
\SetAlgoLined
{\bf Input}: Target data $D_T$, source data $D_S$, privacy data $D_{priv}$, hyperparameter $\lambda$ \\
{\bf Output}: PrivNet \\
\For{number of training iterations}{
1. Accumulate (user, attributes) with a mini-batch $(\mathcal{U}_b, \mathcal{Y}_{b})$ from $D_{priv}$

2. Feed users $\mathcal{U}_b$ and their history in $D_S$ \\ into the source network (see Sec.~\ref{paper:rec}) \\ so as to generate  the transferred \\ knowledge $\mathcal{X}_{b|S}$

3. Update $\Theta$ using examples $(\mathcal{X}_{b|S}, \mathcal{Y}_{b})$ \\ via gradient descent over $\mathcal{L}(\Theta)$.

4. Update $\theta$ using mini-batch examples \\ from $D_S$ and $D_T$ with adversary \\ loss via gradient descent over $\tilde{\mathcal{L}}(\theta)$.
}
The gradient-based updates can use any standard gradient-based learning rule. Deep learning library (e.g., TensorFlow) can automatically calculate gradients.
\caption{Training PrivNet.}
\label{algo:training}
\end{algorithm}

\subsection{Model Learning}\label{paper:learning}

The training process of PrivNet is illustrated in Algorithm~\ref{algo:training}. Lines 1-3 are to optimize the privacy part related parameter, i.e., $\Theta$ in $\mathcal{L}(\Theta)$. On line 1, it creates a mini-batch size examples from data $D_{priv}$. Each example contains a user and their corresponding private attributes $(u, \{y_{u,p}\}_{p=1}^n)$. On line 2, it feeds users and their historical items in the source domain to the source network so as to generate transferred knowledge $\bm{x}_{u|trans}$. On line 3, the transferred knowledge and their corresponding private attributes $(\bm{x}_{u|trans}, \{y_{u,p}\}_{p=1}^n)$ are used to train the privacy attacker component by descending its stochastic gradient using the mini-batch examples: $\nabla_{\Theta} \mathcal{L}(\Theta).$ Line 4 is to optimize the recommender part related parameter, i.e., $\theta$ by descending its stochastic gradient with adversary loss using mini-batch examples:$\nabla_{\theta} \tilde{\mathcal{L}}(\theta).$

\subsection{Complexity Analysis}\label{paper:learning-complexity}

The parameter complexity of PrivNet is the addition of its recommender component and the privacy component. The embedding matrices of the recommender dominate the number of parameters as they vary with the input. As a result, the parameter complexity of PrivNet is $\mathcal{O}(d \cdot (n_S + n_T))$ where $d$ is the embedding dimension, and $n_S$ and $n_T$ are the number of items in the source and target domains respectively.

The learning complexity of PrivNet divides into two parts: the forward prediction and backward parameter update. The forward prediction of PrivNet is the addition of its recommender component and two times of the privacy component since the recommender component needs the loss from the privacy component. The complexity of backward parameter update is the addition of its recommender component and the privacy component since they optimize their own parameters.

\section{Experiments}\label{paper:exp}

In this section, we conduct experiments to evaluate both recommendation performance and privacy protection of PrivNet. 

\subsection{Dataset}\label{paper:data}

We evaluate on the following real-world datasets.

{\it Foursquare (FS)} It is a public available data on user-venue checkins~\cite{yang2019privacy}. The source and target domains are divided by the checkin's time, i.e., dealing with the covariate shift issues where the distribution of the input variables change between the old data and the newly collected one. The private user attribute is Gender.

{\it MovieLens (ML)} It is a public available data on user-movie ratings~\cite{harper2016movielens}. We reserve those ratings over three stars as positive feedbacks. The source and target domains are divided by the movie's release year, i.e., transferring from old movies to the new ones. The private user attributes are Gender and Age. Following~\cite{beigi2020privacy}, we categorize Age into three groups: over-45, under-35, and between 35 and 45.


The statistics are summarized in Table~\ref{table:data} and we can see that all of the datasets have more than 99\% sparsity. It is expected that the transfer learning technique is helpful to alleviate the data sparsity issues in these real-world recommendation services.

\begin{table}[]					
\centering	
\resizebox{0.49\textwidth}{!}{					
\begin{tabular}{c | c | cc | cc | c   }		
\toprule											
{\multirow{2}{*}{Data}} & \multirow{2}{*}{\#user} & \multicolumn{2}{c|}{Target domain}  & \multicolumn{2}{c|}{Source domain} &  {\multirow{2}{*}{\makecell{Private \\ attribute}}}  \\
\cline{3-4}\cline{5-6} 					
\multicolumn{1}{c|}{}  &   &   \#item &  \#rating  &   \#item & \#rating  &   \\		
\midrule
FS &  29,515 & 28,199 & 357,553  & 28,407 & 467,810 & G    \\
\midrule
ML &  5,967  & 2,049  & 274,115  & 1,484  & 299,830  & G, A  \\
\bottomrule
\end{tabular}	
}	
\caption{Statistics of datasets. (G=Gender, A=Age)}
\label{table:data}										
\end{table}

\subsection{ Experimental Setting}\label{paper:protocol}


\subsubsection{Evaluation Metric}

For privacy evaluation, we follow the protocol in~\cite{jia2018attriguard} to randomly sample 80\% of users as the training set and treat the remaining users as the test set. The users in the training set has publicly shown their private information while the users in the test set keep it private. We split a small data from the training set as the validation set where the ratio is train:valid:test=7:1:2. For privacy metrics, we compute Precision, Recall, and F1-score in a weighted\footnote{Note, the weighted F1 values are not necessarily equal to the harmonic mean of the corresponding Precision and Recall values.} way which are suitable for imbalanced data distribution~\cite{fawcett2006introduction}. We report results for each private attribute. We first calculate metrics for each label, and then compute their average weighted by support (the number of true instances for each label). A lower value indicates better privacy protection.

For recommendation evaluation, we follow the leave-one-out strategy in~\cite{he2017neural}, i.e., reserving the latest one interaction as the test item for each user, then randomly sampling a number of (e.g., 99) negative items that are not interacted by the user. We evaluate how well the recommender can rank the test item against these negative ones. We split a small data from the training set as the validation set where the ratio is train:valid:test=7:1:2. For recommendation metrics, we compute hit ratio (HR), normalized discounted cumulative gain (NDCG), mean reciprocal rank (MRR), and AUC for top-$K$ (default $K=10$) item recommendation~\cite{gao2019cross}. A higher value indicates better recommendation.

\begin{table}
\centering
\resizebox{.45\textwidth}{!}{
\begin{tabular}{|c|c|}
\hline
 Hyperparameter & Setting \\
\hline
 train:valid:test & 7:1:2 \\
 user representation size & 80\\
 item representation size & 80\\
 history length cutoff (\#items) & 10 \\
 neural collaborative filtering layers & [80, 64]\\
 attention unit layers & [80, 64]\\
  number of transfer layers & 1\\
 negative sampling ratio for training & 1 \\
 test positive:negative & 1:99 \\
 clip norm & 5\\
 batch size & 128 \\
 bias init & 0 \\
 weight init & Glorot uniform   \\
 embedding init & Glorot uniform  \\
learning rate & 5e-4 \\
 optimizer & Adam \\
 activation function & sigmoid \\
total epochs (with early stopping) & 50\\
\hline
\end{tabular}
}
\caption{Setting of hyperparameters.}
\label{table:hyper}
\end{table}

\subsubsection{Implementation}

All methods are implemented using TensorFlow. Parameters are initialized by default. The optimizer is the adaptive moment estimation with learning rate 5e-4. The size of mini-batch is 128 with negative sampling ratio 1. The embedding size is 80 while the MLP has one hidden layer with size 64. The history size is 10. $\lambda$ is 1 in Eq. (\ref{eq:privnet}). The noise level is 10\%. The number of dummy items are 5. The privacy related metrics are computed by Python scikit-learn library. The setting of hyperparameters used to train our model and the baselines is summarized in Table~\ref{table:hyper}.

\subsection{Baseline}

\begin{table}[]	
\resizebox{0.49\textwidth}{!}{
\begin{tabular}{|l|c|c|}
\hline
 Methods                       & \makecell{Knowledge \\ transfer} & \makecell{Privacy protection \\ (+strategy)} \\
\hline
BPRMF~\cite{BPRMF}               & \xmark             &  \xmark   \\
MLP~\cite{he2017neural}          & \xmark             &  \xmark   \\
CSN~\cite{misra2016cross}        & \cmark             &  \xmark   \\
CoNet~\cite{hu2018conet}         & \cmark             &  \xmark    \\
BlurMe~\cite{weinsberg2012blurme}& \xmark             &  \cmark ~(+perturbation)  \\
LDP~\cite{bassily2015local}      & \xmark             &  \cmark ~(+noise)  \\
PrivNet (ours)                   &  \cmark            &  \cmark ~(+adversary) \\
\hline
\end{tabular}
}
\caption{Categorization of comparing methods. }
\label{table:baselines}										
\end{table}

We compare PrivNet with various kinds of baselines as summarized in Table~\ref{table:baselines}.

The following methods are privacy-agnostic. {\it BPRMF}: Bayesian personalized ranking~\cite{BPRMF} is a latent factors approach which learns user and item factors via matrix factorization. {\it MLP}: Multilayer perceptron~\cite{he2017neural} is a neural CF approach which learns the user-item interaction function using neural networks. {\it CSN}: The cross-stitch network~\cite{misra2016cross} is a deep transfer learning model which couples the two basic networks via a linear combination of activation maps using a translation scalar. {\it CoNet}: Collaborative cross network~\cite{hu2018conet} is a deep transfer learning method for cross-domain recommendation which learns linear combination of activation maps using a translation matrix.

\begin{table*}[t]
\center
\resizebox{0.75\textwidth}{!}{	
\begin{tabular}{c| c| c|c|c|c | c|c| c}
\toprule
Dataset       & Metric & BPRMF & MLP   & CSN   & CoNet & BlurMe & LDP  & PrivNet \\
\midrule
              & HR     & 36.5 & 47.0 & 52.7 & 53.4* & 52.6   & 44.5    & {\bf 54.3}    \\
\cline{2-9}
Foursquare    & NDCG   & 22.0 & 31.5 & 35.9 & 36.3* & 35.4   & 29.9    & {\bf 36.8}    \\
\cline{2-9}
              & MRR    & 17.6 & 31.9 & 35.0* & {\bf 35.3} & 32.1   & 27.1    & 33.4    \\
\midrule
              & HR     & 53.0 & 77.4 & 82.7 & 77.1 & 85.7   & 85.8*    & {\bf 86.0}    \\
\cline{2-9}
MovieLens     & NDCG   & 37.0 & 50.5 & 55.7 & 50.7 & 69.7   & {\bf 69.9}   & {\bf 69.9}    \\
\cline{2-9}
              & MRR    & 32.0 & 44.5 & 49.3 & 44.6 & {\bf 65.9}   & {\bf 65.9}    & 65.7*    \\
\bottomrule
\end{tabular}
}
\caption{Comparison results of different methods on recommendation performance. The bold face indicates the best result while the star mark indicates the second best.}
\label{tb:result-recommender}
\end{table*}

The following methods are privacy-aware. {\it BlurMe}: This method~\cite{weinsberg2012blurme} perturbs a user's profile by adding dummy items to their history. It is a representative of the perturbation-based technique to recommend items while protect private attributes. {\it LDP}: Local differential privacy~\cite{bassily2015local} modifies user-item ratings by adding noise to them based on the differential privacy. It is a representative of the noise-based technique to recommend items while protect private attributes. Note, the original LDP and BlurMe are single-domain models which are also used as comparing baselines in~\cite{beigi2020privacy}. To be fair and to investigate the influence of privacy-preserving strategies, we replace the adversary strategy of PrivNet with the strategy of LDP (adding noise) and BlurMe (perturbing ratings), and keep the other components the same. 

\subsection{Result on Recommendation Performance}\label{paper:result-recommender}

The results of different methods on recommendation are summarized in Table~\ref{tb:result-recommender}. A higher value indicates better recommendation performance.

Comparing with the privacy-agnostic methods (BPRMF, MLP, CSN, and CoNet), PrivNet is superior than them with a large margin on the MovieLens dataset. This shows that PrivNet is effective in recommendation while it protects the source private attributes. Since these four methods represent a wide range of typical recommendation methods (matrix factorization, neural CF, transfer learning), we can see that the architecture of PrivNet is a reasonable design for recommender systems.

Comparing with the privacy-aware methods (LDP and BlurMe), we can see that LDP significantly degrades recommendation performance with a reduction about six to ten percentage points on the Foursquare dataset. This shows that LDP suffers from the noisy source information since it harms the usefulness of the transferred knowledge to the target task. For BlurMe, we can see that BlurMe still degrades recommendation performance on the Foursquare dataset, for example with relative 4.0\% performance reduction in terms of MRR. This shows that BlurMe suffers from the perturbed source information since it harms the usefulness of the transferred knowledge to the target task.

Among the privacy-aware methods, PrivNet achieves the best recommendation performance in terms of all HR, NDCG, and MRR on the Foursquare dataset, and the best in terms of HR on the MovieLens dataset. It shows that PrivNet is better for improving the usefulness of the transferred knowledge by comparing with LDP and BlurMe.

In summary, PrivNet is effective in transferring the knowledge, showing that the adversary strategy of PrivNet achieves state-of-the-art performance by comparing with the strategies of adding noise (LDP) and perturbing ratings (BlurMe).

\begin{table}[t]
\center
\resizebox{0.45\textwidth}{!}{	
\begin{tabular}{c| c| c | c | c}
\toprule
 Dataset            &  Metric  &   LDP   &   BlurMe  &  PrivNet  \\
\midrule
\multirow{3}{*}{Foursquare} & Precision   & {\bf 64.7} & 73.2         & 66.8 \\
                      & Recall      & 75.2       & 75.3         & {\bf 71.1} \\
                      & F1          & {\bf 66.0} & 66.7         & 68.1 \\
\hline
\multirow{3}{*}{MovieLens-G} & Precision   & 73.4       & {\bf 69.4}   & 70.9 \\
                      & Recall      & 75.4       & {\bf 71.7}   & 72.5 \\
                      & F1          & 73.6       & 70.1         & {\bf 62.0} \\
\hline
\multirow{3}{*}{MovieLens-A} & Precision   & 63.8       & {\bf 54.6}  & 55.4 \\
                      & Recall      & 65.5       & 58.1        & {\bf 57.9} \\
                      & F1          & 61.4       & 54.2        & {\bf 46.3} \\
\bottomrule
\end{tabular}
}
\caption{Comparison results on privacy protection. The bold face indicates the best result (the lower the better).}
\label{tb:result-privacy}
\end{table}

\subsection{Result on Privacy Protection}\label{paper:result-privacy}

The results of different methods on privacy inference are summarized in Table~\ref{tb:result-privacy} (Note, there are no results for the four privacy-agnostic methods). A lower value indicates better privacy protection.

Comparing PrivNet and BlurMe, we can see that the perturbation method by adding dummy items still suffers from privacy inference attacks in terms of Precision and Recall on the Foursquare dataset, and in terms of F1 on the MovieLens dataset. The reason may be that the attacker can effectively distinguish the true profiles from the dummy items. That is, it can accurately learn from the true profiles while ignore the dummy items. Comparing PrivNet and LDP, we can see that adding noise to ratings still suffers from privacy inference attacks in terms of Recall on the Foursquare dataset, and in terms of all three metrics on the MovieLens dataset. It implies that the occurrence of a rating, regardless of its numeric value (true or noisy), leaks the user privacy. That is, the binary event of excluding or including an item in a user's profile is a signal for user privacy inference nearly as strong as numerical ratings. In particular, there are 50 movies rated by Female only (e.g., Country Life (1994)) while 350 by Male only (e.g., Time Masters (1982)). Adding noise to these ratings may not influence the inference of Gender for these users very much.

\begin{table*}[]
\center
\resizebox{0.82\textwidth}{!}{	
\begin{tabular}{c| ccc | ccc | ccc}
\toprule
\multirow{2}{*}{Adversary loss?}  & \multicolumn{3}{c|}{Foursquare} & \multicolumn{3}{c|}{MovieLens-Gender} & \multicolumn{3}{c}{MovieLens-Age} \\
 & Precision & Recall    & F1    & Precision   & Recall   & F1  & Precision   & Recall   & F1  \\
\midrule
No              & 73.2   & 75.4  & 68.5  & 73.6   & 75.6   & 73.5   & 60.8   & 65.3   & 61.4   \\
\hline
Yes & {\bf 66.8} & {\bf 71.1} & {\bf 68.1} & {\bf 70.9} & {\bf 72.5} & {\bf 62.0} & {\bf 55.4} & {\bf 57.9} & {\bf 46.3} \\
\bottomrule
\end{tabular}
}
\caption{Necessity of adversary loss to regularize the recommender (lower value better privacy protection).}
\label{tb:necessity-adversary-loss}
\end{table*}

PrivNet achieves nearly half the best results on privacy protection in terms of three evaluation metrics on the two datasets. It has significantly lower F1 scores in comparison to all baselines on the MovieLens dataset. It is effective to hide private information during the knowledge transfer. By simulating the attacks during the training, PrivNet is prepared against the malicious attacks for unseen users in the future. In summary, PrivNet is an effective source privacy-aware transfer model such that it makes the malicious attackers more difficult to infer the source user privacy during the knowledge transfer, compared with the strategies of adding noise (LDP) and perturbing ratings (BlurMe).

\subsubsection{Clustering}\label{paper:result-clustering}
Figure~\ref{fig:tsne} shows t-SNE projections of 4,726 users' transferred representations on the MovieLens-Gender dataset. These user vectors are computed from the user encoder as shown in Figure~\ref{fig:architecture}. We can see that the vectors are more mixed distributed among male and female users with the training of PrivNet. In contract, the vectors for female users are clustered on the top-left corner while male users are on the bottom-right without the training of PrivNet ($\lambda=0$, see Section~\ref{paper:model-ablation}). To quantify the difference, we perform K-means clustering on the user vectors where K=2, and calculate the V-measure~\cite{rosenberg2007v} which assesses the degree of overlap between the 2 clusters and the Gender groups. The measure is 0.0119 and 0.0027 respectively for without and with training of PrivNet. Note that a lower measure is better since we do not want to the two classes to be easily separable.

\subsection{Parameter Sensitivity}\label{paper:analyses}

In this section, we analyse the model ablation, impact of privacy inference component, and impact of public users who share their profiles.

\subsubsection{Model Ablation}\label{paper:model-ablation}
The key component of PrivNet is the adversary loss used to regularize the recommender. We remove this component to show its necessity to protect the private attributes by setting the $\lambda=0$ in Eq. (\ref{eq:privnet}). The results are summarized in Table~\ref{tb:necessity-adversary-loss}. As we expect, PrivNet without adversary loss is most vulnerable to privacy attacks since it has no privacy defense. There is a significant drop in terms of all three privacy-related metrics without this model component.

\begin{figure}
\centering
\begin{subfigure}{.23\textwidth}
\includegraphics[height=1.27in,width=3.75cm]{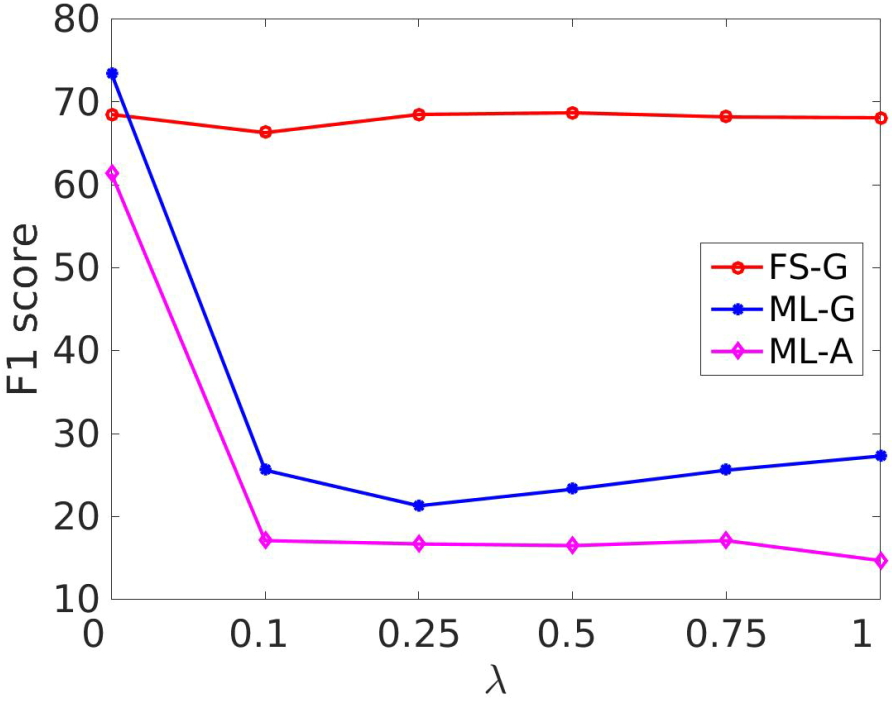}
\caption{Privacy attacker.}
\label{fig:f1-lambda}
\end{subfigure}
\begin{subfigure}{.23\textwidth}
  \includegraphics[height=1.27in,width=3.8cm]{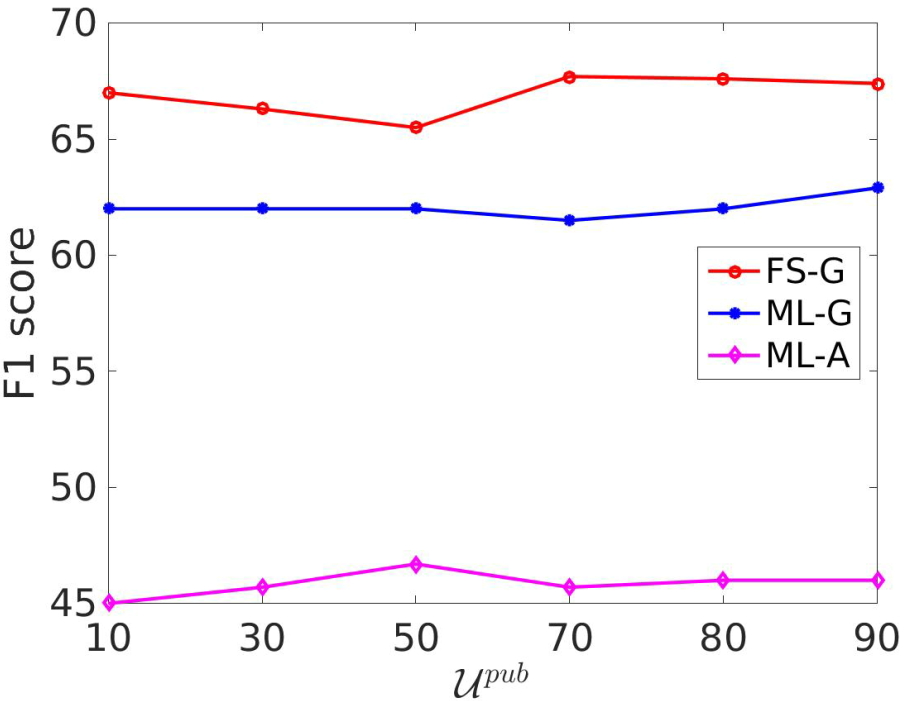}
\caption{Public users.}
\label{fig:f1-public}
\end{subfigure}
\caption{Impact of privacy component and public users. (FS-G: Foursquare-Gender, ML-G: MovieLens-Gender, ML-A: MovieLens-Age)}
\label{fig:sensitivity}
\end{figure}

\subsubsection{Impact of Privacy Component}\label{paper:hyperparameter}

We vary the $\lambda$ (see Eq. (\ref{eq:privnet})) of privacy component with $\{0, 0.1, 0.25, 0.5, 0.75, 1.0\}$ to show the its impact on privacy protection and recommendation (where $\lambda=0$ corresponds to without privacy attack component, see also Table~\ref{tb:necessity-adversary-loss}). Figure~\ref{fig:f1-lambda} shows the impact on privacy protection. The privacy inference generally becomes more difficult with the increase of $\lambda$, showing that the privacy inference component of PrivNet is a key factor for protecting the user privacy in the source domain. In particular, all results of $\lambda \neq 0$ are better than that of $\lambda=0$ in hiding the private information. Privacy inference results, however, are subtle among different private attributes and evaluation metrics. On the Foursquare dataset, F1 decreases at first (until $\lambda$ to 0.1), then it increases. On the MovieLens-Gender dataset, the F1 score decreases at first (until $\lambda$ to 0.25) and then it increases. It means that the private information is obscured more successfully in the beginning but less in the end. The reason may be that the model overfits by increasing the value of $\lambda$ and leads to an inaccurate estimation of privacy inference. On the MovieLens-Age dataset, the F1 score consistently decreases with the increase of $\lambda$.

\begin{figure}
\centering
\begin{subfigure}{.24\textwidth}
  \includegraphics[height=1.5in,width=3.8cm]{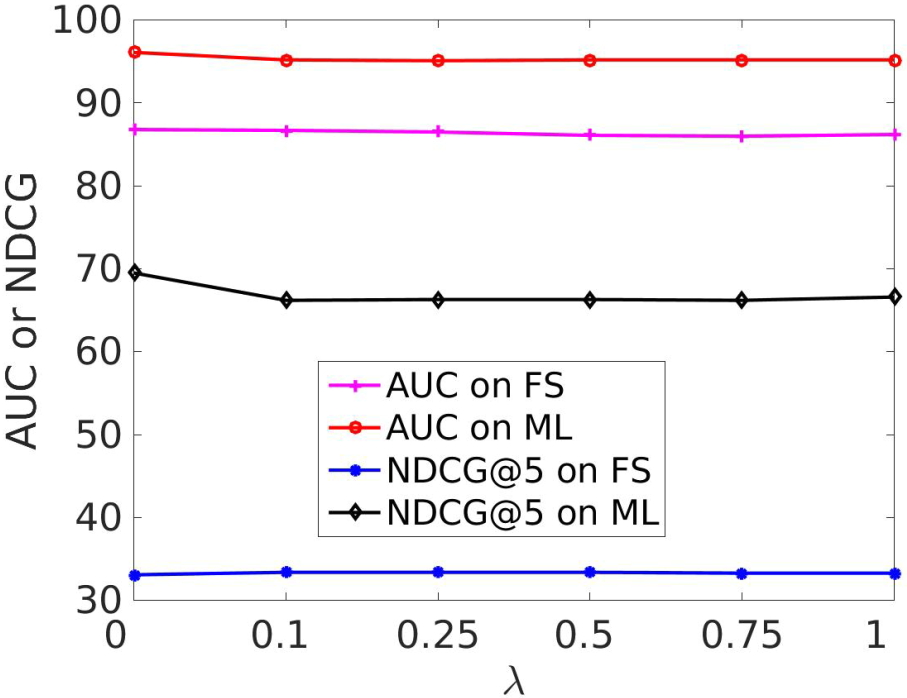}
\caption{Privacy component}
\label{fig:rec-lambda}
\end{subfigure}
\begin{subfigure}{.23\textwidth}
  \includegraphics[height=1.5in,width=3.8cm]{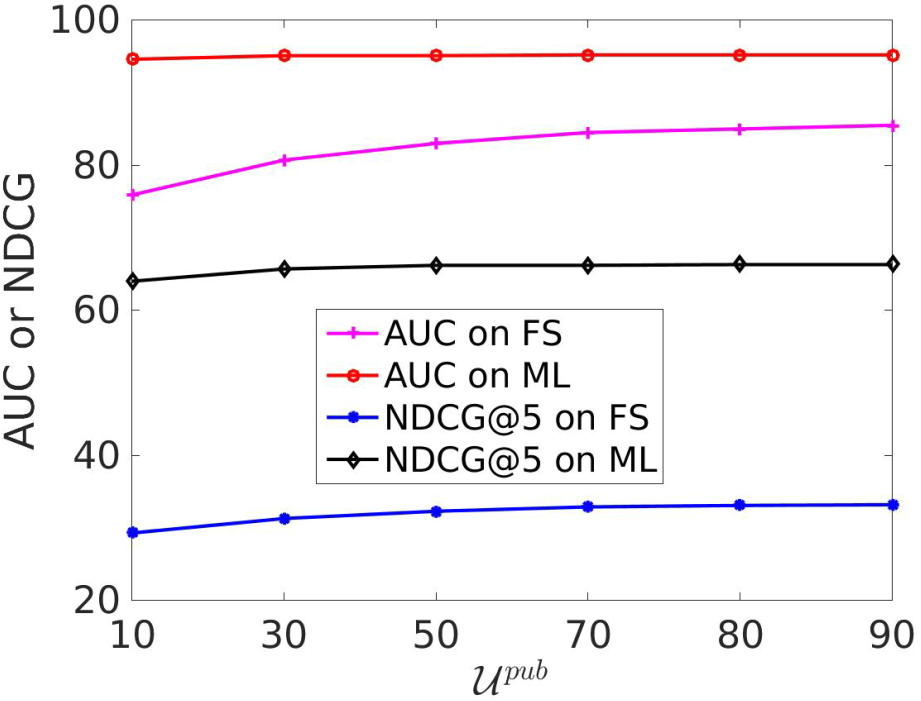}
\caption{Public users}
\label{fig:rec-public}
\end{subfigure}
\caption{Parameter sensitivity for recommendation.}
\label{fig:public}
\end{figure}

Figure~\ref{fig:rec-lambda} shows the impact on recommendation performance. The recommendation performance decreases with $\lambda$ increasing from 0 to 0.1 on the MovieLens dataset, showing that increasing the impact of privacy inference component harms the recommendation quality to some extent.

\subsubsection{Impact of Public Users}\label{paper:public-user}

We vary the percentage of public users $\mathcal{U}^{pub}$ (see Section~\ref{paper:statement}) with $\{10,30,50,70,80,90\}$. Figure~\ref{fig:f1-public} shows the impact on the privacy inference. It is surprising that the privacy inference does not become more easy with the increase of public users. On the Foursquare dataset, it infers inaccurately until the percentage increases to 50\% and then accurately until to 80\% in terms of F1. This shows that the adversary strategy of PrivNet is effective to protect unseen users' privacy when only a small number of users (e.g., 10\%) reveal their profiles for the training. On the MovieLens dataset, it infers inaccurately after 50\% until to 80\% in terms of F1.

Figure~\ref{fig:rec-public} shows the impact on recommendation performance. Since the amount of public users controls how much knowledge is shared between the source and target domains, the recommendation performance improves with the increasing amount of public users. In summary, PrivNet is favourable in practice since it can achieve a good tradeoff on the utility and privacy when only a small amount of users reveal their profiles to the public.

\begin{table}[]
\centering
\resizebox{0.5\textwidth}{!}{
\begin{tabular}{c | c | c| c }
\toprule
{\bf No.} & {\bf Movie} & {\bf Genre} & {\bf Attn weight} \\
\midrule
0  & Chicken Run        & Animation, Children, Comedy & {\it 0.127} \\
\hline
1  & X-Men            & Action, Sci-Fi & 0.069\\
\hline
2  & Mission: Impossible   & Action, Adventure, Mystery & 0.001\\
\hline
3  & Titan A.E.          & Adventure, Animation, Sci-Fi &  0.059\\
\hline
4  & The Perfect Storm    & Action, Adventure, Thriller & 0.056\\
\hline
5  & Gone in 60 Seconds    & Action, Crime &0.053\\
\hline
6  & Schindler's List      & Drama, War & 0.098\\
\hline
7  & The Shawshank Redemption  & Drama & {\bf 0.331} \\
\hline
8  & The Matrix        & Action, Sci-Fi, Thriller & 0.062\\
\hline
9  & Shakespeare in Love   & Comedy, Romance & {\it 0.140}\\
\midrule
10 & Howards End      & Drama & N/A \\
\bottomrule
\end{tabular}
}
\caption{Example: Capturing short-/long-term user interests and high-level
category relationship among items.}
\label{tb:case-study-1}
\end{table}

\subsection{Case Study}\label{paper:case-study}

One advantage of PrivNet is that it can explain which item in a user's history matters the most for a candidate item by using the attention weights. Table~\ref{tb:case-study-1} shows an example of interactions between a user's historical movies (No. 0$\sim$9) and the candidate movie (No. 10). We can see that the latest movie matters a lot since the user interests may remain the same during a short period. The oldest movie, however, also has some impact on the candidate movie, reflecting that the user interests may mix with a long-term characteristic. PrivNet can capture these subtle short-/long-term user interests. Furthermore, the movie (No. 7) belonging to the same genre as the candidate movie matters the most. PrivNet can also capture this high-level category relationship.

\section{Conclusion}\label{paper:conclusion}

We presented an attack scenario to infer the private user attributes from the transferred knowledge in recommendation, raising the issues of source privacy leakage beyond target performance. To protect user privacy in the source domain, a privacy-aware transfer model (PrivNet) is proposed beyond improving the performance in the target domain. It is effective in terms of recommendation performance and privacy protection, achieving a good trade-off between the utility and privacy of the transferred knowledge. In future works, we want to relax the assumption that the private user attributes need to provide in advance in order to train the privacy inference component for protecting unseen users.

\section*{Acknowledgement}
We thank Dr. Yu Zhang for insightful discussion. We thank the new publication paradigm, i.e., ``Findings of ACL: EMNLP 2020'', which makes this paper indexed in the ACL anthology. The work was supported by Hong Kong CERG projects 16209715/16244616, and Hong Kong PhD Fellowship Scheme.

\bibliography{emnlp2020}

\begin{thebibliography}{56}
\expandafter\ifx\csname natexlab\endcsname\relax\def\natexlab#1{#1}\fi

\bibitem[{Bassily and Smith(2015)}]{bassily2015local}
R.~Bassily and A.~Smith. 2015.
\newblock Local, private, efficient protocols for succinct histograms.
\newblock In \emph{ACM STOC}.

\bibitem[{Beigi et~al.(2020)Beigi, Mosallanezhad, Guo, Alvari
  et~al.}]{beigi2020privacy}
G.~Beigi, A.~Mosallanezhad, R.~Guo, H.~Alvari, et~al. 2020.
\newblock Privacy-aware recommendation with private-attribute protection using
  adversarial learning.
\newblock In \emph{ACM WSDM}.

\bibitem[{Cantador et~al.(2015)Cantador, Fern{\'a}ndez-Tob{\'\i}as, Berkovsky,
  and Cremonesi}]{cantador2015cross}
Iv{\'a}n Cantador, Ignacio Fern{\'a}ndez-Tob{\'\i}as, Shlomo Berkovsky, and
  Paolo Cremonesi. 2015.
\newblock Cross-domain recommender systems.
\newblock In \emph{Recommender systems handbook}, pages 919--959.

\bibitem[{Chen et~al.(2019)Chen, Zhang, Wang, Ma, Li, Liu, and
  Ma}]{Chen2019EAT}
Chong Chen, Min Zhang, Chenyang Wang, Weizhi Ma, Minming Li, Yiqun Liu, and
  Shaoping Ma. 2019.
\newblock An efficient adaptive transfer neural network for social-aware
  recommendation.
\newblock In \emph{ACM SIGIR}.

\bibitem[{Chen et~al.(2018)Chen, Dong, Li, and He}]{chen2018federated}
F.~Chen, Z.~Dong, Z.~Li, and X.~He. 2018.
\newblock Federated meta-learning for recommendation.
\newblock \emph{arXiv:1802.07876}.

\bibitem[{Cheng et~al.(2016)Cheng, Koc, Harmsen, Shaked et~al.}]{cheng2016wide}
H.~Cheng, L.~Koc, J.~Harmsen, T.~Shaked, et~al. 2016.
\newblock Wide \& deep learning for recommender systems.
\newblock In \emph{ACM RecSys Workshop}.

\bibitem[{Dwork et~al.(2006)Dwork, McSherry, Nissim, and
  Smith}]{dwork2006calibrating}
C.~Dwork, F.~McSherry, K.~Nissim, and A.~Smith. 2006.
\newblock Calibrating noise to sensitivity in private data analysis.
\newblock In \emph{Theory of Cryptography}.

\bibitem[{Elkahky et~al.(2015)Elkahky, Song, and He}]{elkahky2015multi}
A.~Elkahky, Y.~Song, and X.~He. 2015.
\newblock A multi-view deep learning approach for cross domain user modeling in
  recommendation systems.
\newblock In \emph{WWW}.

\bibitem[{Fawcett(2006)}]{fawcett2006introduction}
T.~Fawcett. 2006.
\newblock An introduction to roc analysis.
\newblock \emph{Pattern recognition letters}.

\bibitem[{Gao et~al.(2019{\natexlab{a}})Gao, Chen, Feng, Zhao
  et~al.}]{gao2019cross}
C.~Gao, X.~Chen, F.~Feng, K.~Zhao, et~al. 2019{\natexlab{a}}.
\newblock Cross-domain recommendation without sharing user-relevant data.
\newblock In \emph{WWW}.

\bibitem[{Gao et~al.(2019{\natexlab{b}})Gao, He, Gan, Chen, Feng, Li, Chua, and
  Jin}]{gao2019neural}
Chen Gao, Xiangnan He, Dahua Gan, Xiangning Chen, Fuli Feng, Yong Li, Tat-Seng
  Chua, and Depeng Jin. 2019{\natexlab{b}}.
\newblock Neural multi-task recommendation from multi-behavior data.
\newblock In \emph{IEEE ICDE}.

\bibitem[{Gao et~al.(2010)Gao, Tian, Huang, and Yang}]{gao2010vlogging}
W.~Gao, Y.~Tian, T.~Huang, and Q.~Yang. 2010.
\newblock Vlogging: A survey of videoblogging technology on the web.
\newblock \emph{ACM Computing Surveys}.

\bibitem[{Goodfellow et~al.(2014)Goodfellow, Pouget, Mirza, Xu
  et~al.}]{goodfellow2014generative}
I.~Goodfellow, J.~Pouget, M.~Mirza, B.~Xu, et~al. 2014.
\newblock Generative adversarial nets.
\newblock In \emph{NIPS}.

\bibitem[{Harper and Konstan(2016)}]{harper2016movielens}
F.~Harper and J.~Konstan. 2016.
\newblock The movielens datasets: History and context.
\newblock \emph{ACM TIST}.

\bibitem[{He et~al.(2017)He, Liao, Zhang, Nie et~al.}]{he2017neural}
X.~He, L.~Liao, H.~Zhang, L.~Nie, et~al. 2017.
\newblock Neural collaborative filtering.
\newblock In \emph{WWW}.

\bibitem[{Hu et~al.(2018)Hu, Zhang, and Yang}]{hu2018conet}
G.~Hu, Y.~Zhang, and Q.~Yang. 2018.
\newblock Conet: Collaborative cross networks for cross-domain recommendation.
\newblock In \emph{ACM CIKM}.

\bibitem[{Hu et~al.(2019)Hu, Zhang, and Yang}]{TMH}
Guangneng Hu, Yu~Zhang, and Qiang Yang. 2019.
\newblock Transfer meets hybrid: a synthetic approach for cross-domain
  collaborative filtering with text.
\newblock In \emph{The World Wide Web Conference}, pages 2822--2829.

\bibitem[{Huang et~al.(2016)Huang, Zhang, Gong, and Huang}]{huang2016hashtag}
H.~Huang, Q.~Zhang, Y.~Gong, and X.~Huang. 2016.
\newblock Hashtag recommendation using end-to-end memory networks with
  hierarchical attention.
\newblock In \emph{COLING}.

\bibitem[{Huang and Lin(2016)}]{huang2016transferring}
Y.~Huang and S.~Lin. 2016.
\newblock Transferring user interests across websites with unstructured text
  for cold-start recommendation.
\newblock In \emph{EMNLP}.

\bibitem[{Jia and Gong(2018)}]{jia2018attriguard}
J.~Jia and N.~Gong. 2018.
\newblock Attriguard: A practical defense against attribute inference attacks
  via adversarial machine learning.
\newblock In \emph{USENIX Security}.

\bibitem[{Li et~al.(2009)Li, Yang, and Xue}]{li2009can}
B.~Li, Q.~Yang, and X.~Xue. 2009.
\newblock Can movies and books collaborate? cross-domain collaborative
  filtering for sparsity reduction.
\newblock In \emph{IJCAI}.

\bibitem[{Liu et~al.(2018)Liu, Wei, Zhang, Yan, and Yang}]{liu2018transferable}
B.~Liu, Y.~Wei, Y.~Zhang, Z.~Yan, and Q.~Yang. 2018.
\newblock Transferable contextual bandit for cross-domain recommendation.
\newblock In \emph{AAAI}.

\bibitem[{Ma et~al.(2019{\natexlab{a}})Ma, Ren, Lin, Chen, Ma, and
  Rijke}]{Ma2019PI}
Muyang Ma, Pengjie Ren, Yujie Lin, Zhumin Chen, Jun Ma, and Maarten~de Rijke.
  2019{\natexlab{a}}.
\newblock Pi-net: A parallel information-sharing network for shared-account
  cross-domain sequential recommendations.
\newblock In \emph{ACM SIGIR}.

\bibitem[{Ma et~al.(2019{\natexlab{b}})Ma, Zong, Yang, and Su}]{ma2019news2vec}
Y.~Ma, L.~Zong, Y.~Yang, and J.~Su. 2019{\natexlab{b}}.
\newblock News2vec: News network embedding with subnode information.
\newblock In \emph{EMNLP}.

\bibitem[{Man et~al.(2017)Man, Shen, Jin, and Cheng}]{man2017cross}
Tong Man, Huawei Shen, Xiaolong Jin, and Xueqi Cheng. 2017.
\newblock Cross-domain recommendation: an embedding and mapping approach.
\newblock In \emph{IJCAI}.

\bibitem[{McSherry and Mironov(2009)}]{mcsherry2009differentially}
F.~McSherry and I.~Mironov. 2009.
\newblock Differentially private recommender systems: Building privacy into the
  netflix prize contenders.
\newblock In \emph{ACM SIGKDD}.

\bibitem[{Meng et~al.(2018)Meng, Wang, Shu, Li et~al.}]{meng2018personalized}
X.~Meng, S.~Wang, K.~Shu, J.~Li, et~al. 2018.
\newblock Personalized privacy-preserving social recommendation.
\newblock In \emph{AAAI}.

\bibitem[{Misra et~al.(2016)Misra, Shrivastava, Gupta, and
  Hebert}]{misra2016cross}
Ishan Misra, Abhinav Shrivastava, Abhinav Gupta, and Martial Hebert. 2016.
\newblock Cross-stitch networks for multi-task learning.
\newblock In \emph{IEEE CVPR}.

\bibitem[{Nikolaenko et~al.(2013)Nikolaenko, Ioannidis, Weinsberg, Joye
  et~al.}]{nikolaenko2013privacy}
V.~Nikolaenko, S.~Ioannidis, U.~Weinsberg, M.~Joye, et~al. 2013.
\newblock Privacy-preserving matrix factorization.
\newblock In \emph{ACM CCS}.

\bibitem[{Oquab et~al.(2014)Oquab, Bottou, Laptev, and
  Sivic}]{oquab2014learning}
Maxime Oquab, Leon Bottou, Ivan Laptev, and Josef Sivic. 2014.
\newblock Learning and transferring mid-level image representations using
  convolutional neural networks.
\newblock In \emph{Proceedings of the IEEE conference on computer vision and
  pattern recognition}, pages 1717--1724.

\bibitem[{Pan et~al.(2008)Pan, Zhou, Cao, Liu et~al.}]{pan2008one}
R.~Pan, Y.~Zhou, B.~Cao, N.~Liu, et~al. 2008.
\newblock One-class collaborative filtering.
\newblock In \emph{IEEE ICDM}.

\bibitem[{Pan and Yang(2009)}]{pan2009survey}
Sinno~Jialin Pan and Qiang Yang. 2009.
\newblock A survey on transfer learning.
\newblock \emph{IEEE Transactions on knowledge and data engineering},
  22(10):1345--1359.

\bibitem[{Pan et~al.(2010)Pan, Xiang, Liu, and Yang}]{pan2010transfer}
W.~Pan, E.~Xiang, N.~Liu, and Q.~Yang. 2010.
\newblock Transfer learning in collaborative filtering for sparsity reduction.
\newblock In \emph{AAAI}.

\bibitem[{Polat and Du(2003)}]{polat2003privacy}
H.~Polat and W.~Du. 2003.
\newblock Privacy-preserving collaborative filtering using randomized
  perturbation techniques.
\newblock In \emph{IEEE ICDM}.

\bibitem[{Ramakrishnan et~al.(2001)Ramakrishnan, Keller, Mirza, Grama, and
  Karypis}]{ramakrishnan2001privacy}
N.~Ramakrishnan, B.~Keller, B.~Mirza, A.~Grama, and G.~Karypis. 2001.
\newblock Privacy risks in recommender systems.
\newblock \emph{IEEE Internet Computing}.

\bibitem[{Ravfogel et~al.(2020)Ravfogel, Elazar, Gonen, Twiton, and
  Goldberg}]{ravfogel2020null}
Shauli Ravfogel, Yanai Elazar, Hila Gonen, Michael Twiton, and Yoav Goldberg.
  2020.
\newblock Null it out: Guarding protected attributes by iterative nullspace
  projection.
\newblock \emph{ACL}.

\bibitem[{Rendle et~al.(2009)Rendle, Freudenthaler, Gantner, and
  Schmidt-Thieme}]{BPRMF}
S.~Rendle, C.~Freudenthaler, Z.~Gantner, and L.~Schmidt-Thieme. 2009.
\newblock Bpr: Bayesian personalized ranking from implicit feedback.
\newblock In \emph{UAI}.

\bibitem[{Resheff et~al.(2019)Resheff, Elazar, Shahar, and
  Shalom}]{Resheff19Privacy}
Yehezkel Resheff, Yanai Elazar, Moni Shahar, and Oren Shalom. 2019.
\newblock Privacy and fairness in recommender systems via adversarial training
  of user representations.
\newblock In \emph{Proceedings of the 8th International Conference on Pattern
  Recognition Applications and Methods}.

\bibitem[{Rosenberg and Hirschberg(2007)}]{rosenberg2007v}
Andrew Rosenberg and Julia Hirschberg. 2007.
\newblock V-measure: A conditional entropy-based external cluster evaluation
  measure.
\newblock In \emph{EMNLP}.

\bibitem[{Singh and Gordon(2008)}]{singh2008relational}
Ajit~P Singh and Geoffrey~J Gordon. 2008.
\newblock Relational learning via collective matrix factorization.
\newblock In \emph{SIGKDD}.

\bibitem[{Wan et~al.(2020)Wan, Ni, Misra, and McAuley}]{wan2020addressing}
M.~Wan, J.~Ni, R.~Misra, and J.~McAuley. 2020.
\newblock Addressing marketing bias in product recommendations.
\newblock In \emph{ACM WSDM}.

\bibitem[{Wang et~al.(2018{\natexlab{a}})Wang, Zhang, Xie, and
  Guo}]{wang2018dkn}
H.~Wang, F.~Zhang, X.~Xie, and M.~Guo. 2018{\natexlab{a}}.
\newblock Dkn: Deep knowledge-aware network for news recommendation.
\newblock In \emph{WWW}.

\bibitem[{Wang et~al.(2019)Wang, Tang, Arriaga, and Ryan}]{ijcai2019novel}
J.~Wang, Q.~Tang, A.~Arriaga, and P.~Ryan. 2019.
\newblock Novel collaborative filtering recommender friendly to privacy
  protection.
\newblock In \emph{IJCAI}.

\bibitem[{Wang and Zhou(2020)}]{wang2020pifferentially}
J.~Wang and Z.~Zhou. 2020.
\newblock Differentially private learning with small public data.
\newblock In \emph{AAAI}.

\bibitem[{Wang et~al.(2018{\natexlab{b}})Wang, Gu, and
  Brown}]{wang2018differentially}
Y.~Wang, Q.~Gu, and D.~Brown. 2018{\natexlab{b}}.
\newblock Differentially private hypothesis transfer learning.
\newblock In \emph{ECML-PKDD}.

\bibitem[{Weinsberg et~al.(2012)Weinsberg, Bhagat, Ioannidis, and
  Taft}]{weinsberg2012blurme}
U.~Weinsberg, S.~Bhagat, S.~Ioannidis, and N.~Taft. 2012.
\newblock Blurme: Inferring and obfuscating user gender based on ratings.
\newblock In \emph{ACM RecSys}.

\bibitem[{Yang et~al.(2017{\natexlab{a}})Yang, Bai, Zhang, Yuan, and
  Han}]{yang2017bridging}
Carl Yang, Lanxiao Bai, Chao Zhang, Quan Yuan, and Jiawei Han.
  2017{\natexlab{a}}.
\newblock Bridging collaborative filtering and semi-supervised learning: a
  neural approach for poi recommendation.
\newblock In \emph{ACM SIGKDD}.

\bibitem[{Yang et~al.(2017{\natexlab{b}})Yang, Yan, Yu, Li, and
  Chiu}]{Yang2017MUB}
Chunfeng Yang, Huan Yan, Donghan Yu, Yong Li, and Dah~Ming Chiu.
  2017{\natexlab{b}}.
\newblock Multi-site user behavior modeling and its application in video
  recommendation.
\newblock In \emph{ACM SIGIR}.

\bibitem[{Yang et~al.(2019{\natexlab{a}})Yang, Qu, and
  Cudr{\'e}}]{yang2019privacy}
D.~Yang, B.~Qu, and P.~Cudr{\'e}. 2019{\natexlab{a}}.
\newblock Privacy-preserving social media data publishing for personalized
  ranking-based recommendation.
\newblock \emph{IEEE TKDE}.

\bibitem[{Yang et~al.(2019{\natexlab{b}})Yang, Liu, Chen, and
  Tong}]{yang2019federated}
Q.~Yang, Y.~Liu, T.~Chen, and Y.~Tong. 2019{\natexlab{b}}.
\newblock Federated machine learning: Concept and applications.
\newblock \emph{ACM TIST}.

\bibitem[{Yosinski et~al.(2014)Yosinski, Clune, Bengio, and
  Lipson}]{yosinski2014transferable}
Jason Yosinski, Jeff Clune, Yoshua Bengio, and Hod Lipson. 2014.
\newblock How transferable are features in deep neural networks?
\newblock In \emph{Advances in neural information processing systems}, pages
  3320--3328.

\bibitem[{Yuan et~al.(2019)Yuan, Yao, and Benatallah}]{Yuan2019darec}
Feng Yuan, Lina Yao, and Boualem Benatallah. 2019.
\newblock Darec: Deep domain adaptation for cross-domain recommendation via
  transferring rating patterns.
\newblock In \emph{IJCAI}.

\bibitem[{Zhang et~al.(2016)Zhang, Yuan, Lian, Xie, and
  Ma}]{zhang2016collaborative}
Fuzheng Zhang, Nicholas~Jing Yuan, Defu Lian, Xing Xie, and Wei-Ying Ma. 2016.
\newblock Collaborative knowledge base embedding for recommender systems.
\newblock In \emph{ACM SIGKDD}.

\bibitem[{Zhang and Yang(2017)}]{zhang2017survey}
Yu~Zhang and Qiang Yang. 2017.
\newblock A survey on multi-task learning.
\newblock \emph{arXiv preprint arXiv:1707.08114}.

\bibitem[{Zhao et~al.(2014)Zhao, Guo, He, Jiang et~al.}]{zhao2014we}
X.~Zhao, Y.~Guo, Y.~He, H.~Jiang, et~al. 2014.
\newblock We know what you want to buy: a demographic-based system for product
  recommendation on microblogs.
\newblock In \emph{ACM SIGKDD}.

\bibitem[{Zhou et~al.(2018)Zhou, Zhu, Song, Fan et~al.}]{zhou2018deep}
G.~Zhou, X.~Zhu, C.~Song, Y.~Fan, et~al. 2018.
\newblock Deep interest network for click-through rate prediction.
\newblock In \emph{ACM SIGKDD}.

\end{thebibliography}
\bibliographystyle{acl_natbib}

\end{document}